\documentclass{easychair}

\usepackage{doc}
\usepackage{subcaption}

\title{Data-Driven Construction of Data Center Graph of Things for Anomaly Detection}

\author{
Hao Zhang\inst{1}\thanks{Corresponding author, work primarily done when the author was an intern at Alibaba}
\and
Zhan Li\inst{2}
\and
Zhixing Ren\inst{2}
}

\institute{
Mechanical and Aerospace Engineering, Princeton University, Princeton, NJ 08544, U.S.A.\\
\email{haozhang@princeton.edu}
\and
Alibaba Group Inc. 500 108th Ave NE Suite 800, Bellevue, WA 98004, U.S.A.\\
   \email{\{zhan.li, zhixing.ren\}@alibaba-inc.com}\\
 }

\authorrunning{Zhang, Li and Ren}

\titlerunning{Data Center Graph of Things}

\begin{document}

\maketitle

\begin{abstract}
Data center (DC) contains both IT devices and facility equipment, and the operation of a DC requires a high-quality monitoring (anomaly detection) system. There are lots of sensors in computer rooms for the DC monitoring system, and they are inherently related. This work proposes a data-driven pipeline (\textit{ts2graph}) to build a DC graph of things (sensor graph) from the time series measurements of sensors. The sensor graph is an undirected weighted property graph, where sensors are the nodes, sensor features are the node properties, and sensor connections are the edges. The sensor node property is defined by features that characterize the sensor events (behaviors), instead of the original time series. The sensor connection (edge weight) is defined by the probability of concurrent events between two sensors. A graph of things prototype is constructed from the sensor time series of a real data center, and it successfully reveals meaningful relationships between the sensors. To demonstrate the use of the DC sensor graph for anomaly detection, we compare the performance of graph neural network (GNN) and existing standard methods on synthetic anomaly data. GNN outperforms existing algorithms by a factor of 2 to 3 (in terms of precision and F1 score), because it takes into account the topology relationship between DC sensors. We expect that the DC sensor graph can serve as the infrastructure for the DC monitoring system since it represents the sensor relationships.
\end{abstract}


\setcounter{tocdepth}{2}
{\small
\tableofcontents}

%
%

\section{Introduction}
\label{sec:introduction}
\begin{figure}[th]
    \begin{centering}
    \includegraphics[width=0.5\textwidth]{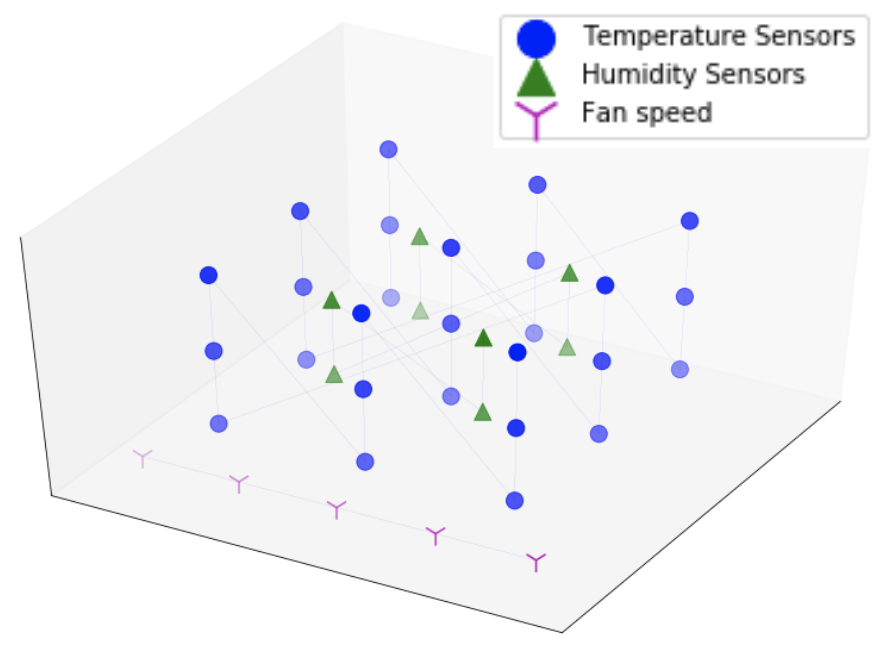}
    \caption{A sketch of sensors in a data center}
    \label{fig:sensor}
    \end{centering}
\end{figure}

Data centers (DC) are the fundamental building blocks of cloud computing~\cite{mell2011nist}. DC holds both IT devices (e.g., computer servers) and facility equipment (e.g., air conditioning). A monitoring system is essential for the operation of DC~\cite{wang2011statistical}. There are lots of sensors in the computer rooms, such as IT loads, temperature, and cooling fan speed. These sensors are crucial devices to monitor the health and performance of the data center. A sketch of data center sensors is shown in Figure~\ref{fig:sensor}.

The sensors in a data center are inherently related to each other. For example, the increase of the temperature could be caused by the failure of a fan of a Computer Room Air Handler (CRAH), or the jump of the IT load. However, the raw sensory data is time series and does not reveal the sensor relationships. The relationships are human expert knowledge and experience, and are unknown to DC monitoring systems. In case an event (say a server is broken, or the power supply is down) happens, it strongly depends on the experience and knowledge of the operators to figure out what happened.

To address this challenge, the DC graph of things (sensor graph) is proposed to represent the relations between DC sensors. There have been a few attempts to apply graph database or graph theory to describe time series data \cite{tripathi2017combination,yao2013applying}. To the best of our knowledge, this work is the first to build a graph of things from the DC sensory data. In particular, the sensor graph is designed for the monitoring system of DC. DC sensory data consists of related multivariate time series and does not naturally form a graph. The dynamic nature of the time series makes them more challenging than simple static objects like entities/concepts in knowledge graph~\cite{singhal2012introducing}.

To convert the time series to a graph, DC sensors will be described by their behaviors, not the raw time series. Examples of time series behaviors include spike and dip, mean shift, variance shift, and trend change. These behaviors are referred to as \textit{events}. Events are interpretable, and sparse in time. The sensor graph is an undirected weighted property graph, where the sensor is the node, sensor feature is the node property, and sensor connection is the edge. The connection between two sensors is defined by their concurrent events (events that happen at about the same time). If two sensors share many common events, then they have strong connections. In case an anomaly alarm occurs (say the temperature increases), the sensor graph will provide information about which sensors to reference in order to identify the root cause. Sensor graph based anomaly detection is expected to be more accurate and robust than time series based anomaly detection.

The main contributions of this work are: (1) we propose data-driven pipeline (\textit{ts2graph}) to construct DC graph of things (sensor graph) from raw sensor time series (section~\ref{sec:t2graph}). The sensor graph represents the relationship between sensors; (2) we validate that \textit{ts2graph} is able to reveal meaningful relationships using real DC sensory time series data (section~\ref{sec:validation}); (3) we demonstrate the use of the sensor graph in anomaly detection. Sensor graph based graph neural network (GNN) outperforms existing standard methods by a factor of 2 to 3 in terms of precision and F1 score (section~\ref{sec:dcanomaly}).

\section{Time Series to Graph of Things}
\label{sec:t2graph}
\begin{figure}[th]
\centering
\includegraphics[width=0.85\columnwidth]{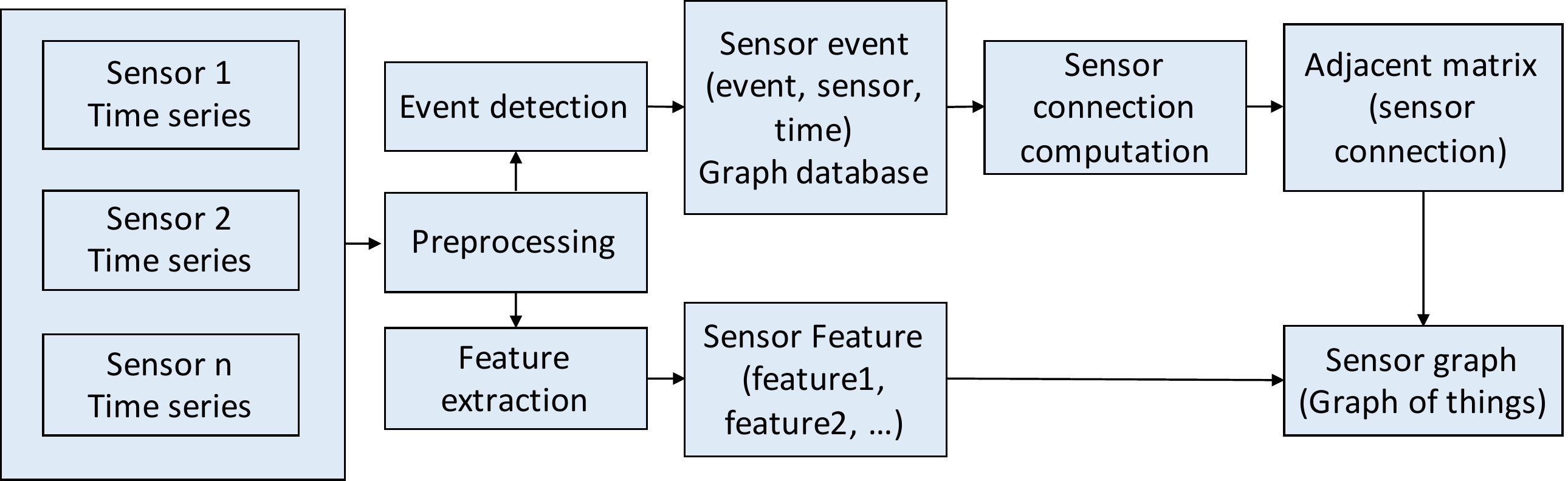} 
\caption{Pipeline (\textit{ts2graph}) to convert time series to graph of things (sensor graph). The raw input is the sensor time series, and the final output is the graph of things (sensor graph).}
\label{fig:ts2graph}
\end{figure}

In this section, we will describe the pipeline (\textit{ts2graph}) to convert a time series to a graph of things (sensor graph), an undirected weighted property graph. From now on, we will use the term ``graph of things" and ``sensor graph" interchangeably. Figure \ref{fig:ts2graph} shows the pipeline for \textit{ts2graph}. All the steps will be explained below.

\subsection{Preprocessing}
The raw time series will be preprocessed, which includes cleaning, normalization, interpolating missing values. Some sensory time series are seasonal. For example, the temperature is affected by the natural cycle of sunrise and sunset, so it has daily seasonality. Therefore, the time-series seasonality must be accounted for before feature extraction and event detection. 

The STL (Seasonal and Trend decomposition using Loess) algorithm \cite{cleveland1990stl} is a standard and robust method for decomposing time series. STL decomposes time series into three components: trend, seasonality, and residual. Specifically, given a time series $x(t)$, it can be decomposed as:
\begin{equation}
x(t) = d(t) + s(t) + r(t),
\end{equation}
where $d(t)$ is the trend, $s(t)$ is the seasonality, and $r(t)$ is the residual. Different event detection requires different components to be used. For example, the trend is used for mean shift, while the residual should be used for variance shift.

\subsection{Feature Extraction}
\begin{figure}[th]
\centering
\includegraphics[width=0.65\columnwidth]{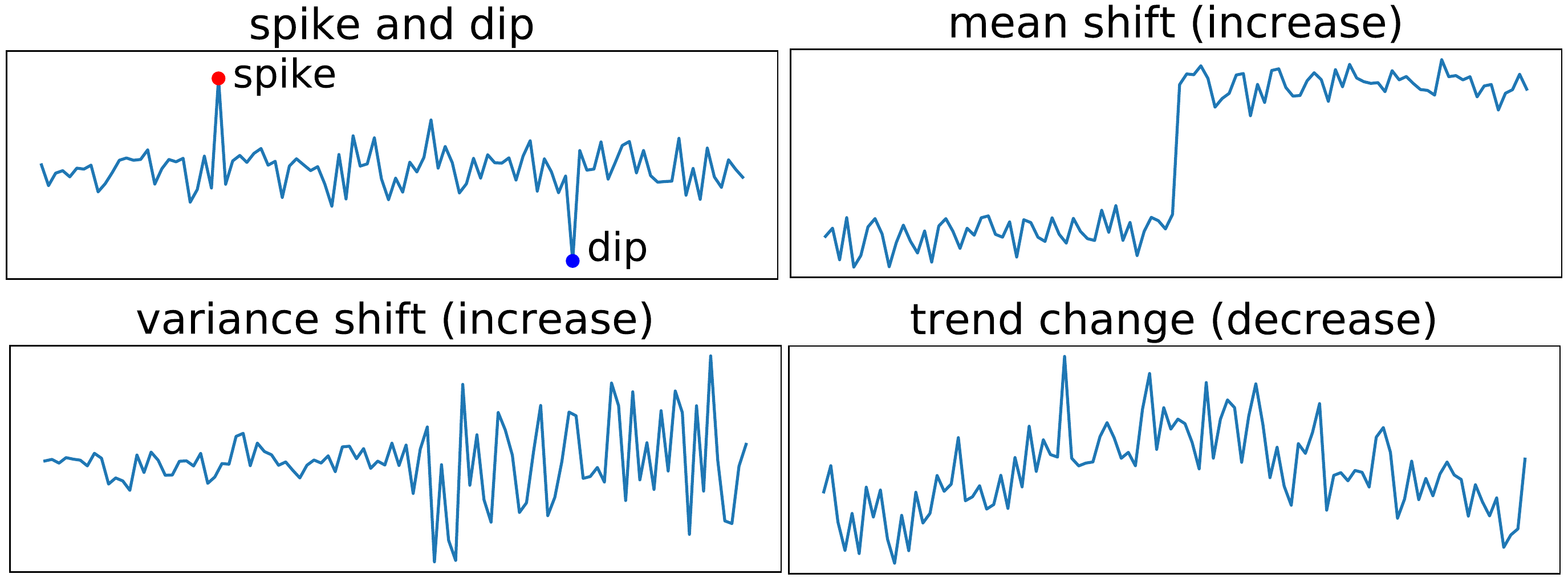} 
\caption{Sketch of four types of events: spike and dip, mean shift, variance shift, and trend change.}
\label{fig:4event}
\end{figure}

The raw time series will be described by behaviors, which we call \textit{events}. Here we consider four types of events: spike and dip, mean shift, variance shift, and trend change. A sketch of these events is shown in Figure \ref{fig:4event}.

The descriptive event definition is intuitive but is not a precise mathematical definition. There exists simple rule like: ``if the temperature increases by \textit{n} degrees in an hour, it is a mean shift event". However, this seemingly simple rule needs extensive manual tuning, which makes it un-robust and not generalizable to different types of sensors and events. Statistical tests are more systematic, such as T-test for equal mean, F-test for equal variance \cite{snedecor1989statistical}, Mann-Kendall test for monotonic trend \cite{mann1945non}. However, due to the non-normality and the non-stationarity, these methods suffer from the same problem: un-robust and hard to tune.

To address this issue, we propose to use the generalized z-score as features to detect events. Roughly speaking, we hope to define a feature for each type of events. Then we will determine if an event happens using its corresponding feature. To start with, we define the z-score transform function for a time series $x(t)$,
\begin{equation}
\zeta(x(t)) = \frac{x(t) - \textnormal{mean}(x(t), T)}{\textnormal{std}(x(t), T)},
\label{eqn:zfunc}
\end{equation}
where $\textnormal{mean}(x(t), T)$, $\textnormal{std}(x(t), T)$ are the mean and standard deviation of $x(t)$ in a recent window of size $T$ respectively. The recent window contains $x(\tau), \tau \in [t-T, t]$. Basically, $\zeta(x(t))$ transforms a time series $x(t)$ to its z-score (also a time series).

The definition of the generalized z-score consists of three steps. First, depending on time series seasonality and event type, we take the appropriate STL component of the time series $x(t)$ (can also be the raw time series if there is no seasonality), denoted as $x_1(t)$. Second, depending on event type, we apply different transformation to $x_1(t)$ to get intermediate feature $x_2(t)$. The last step is to apply the z-score transform to $x_2(t)$ to obtain the corresponding z-score.

\begin{table}[th]
\smallskip
\centering
\resizebox{.8\columnwidth}{!}{
\smallskip
\def\arraystretch{1.5}
\begin{tabular}{|c||c|c|c|}
\hline
        & $x_1(t)$ & $x_2(t)$ & $z(t)$ \\ \hline \hline
Spike and dip & $r(t)$ \vline \ $x(t)$ & $x_1(t)$ & $\zeta(x_2(t))$\\ \hline
Mean shift & $d(t)$ \vline \ $x(t)$ & $\textnormal{mean}(x_1(t), \textnormal{right}) - \textnormal{mean}(x_1(t), \textnormal{left})$ & $\zeta(x_2(t))$\\ \hline
Variance shift & $r(t)$ \vline \ $x(t)$ & $\textnormal{std}(x_1(t), \textnormal{right}) - \textnormal{std}(x_1(t), \textnormal{left})$ & $\zeta(x_2(t))$\\ \hline
Trend change & $d(t)$ \vline \ $x(t)$ & $\textnormal{ema}(\Delta x_1(t), \textnormal{right}) - \textnormal{ema}(\Delta x_1(t), \textnormal{left})$ & $\zeta(x_2(t))$\\ \hline
\end{tabular}
}
\caption{Definition of z-score for spike and dip, mean shift, variance shift, and trend change. $x_1(t)$ is a STL component for seasonal time series, and it is simply $x(t)$ for nonseasonal data.} 
\label{tab:zscore}
\end{table}

The details of these three steps are summarized in Table \ref{tab:zscore}. For seasonal time series, $x_1(t)$ is the trend or residual (depending on event type). The right window (recent time window) contains data $x(\tau)$, $\tau \in [t-\textnormal{right}, t]$, while the left window (less recent time window) contains data $x(\tau)$, $\tau \in [t-\textnormal{right}-\textnormal{left}, t-\textnormal{right}]$. $\textnormal{mean}(x(t), \textnormal{right}), \textnormal{mean}(x(t), \textnormal{left})$ is the mean of $x(t)$ in the right window and left window respectively. $\textnormal{std}(\cdot,\cdot)$ is the standard deviation. $\Delta x(t) = x(t) - x(t-1)$ is the (first order) difference of time series. $\textnormal{ema}(\cdot,\cdot)$ is the exponential moving average.

\subsection{Event Detection}
Given the definition of features, we are ready to determine if an event occurs. The features are generalized z-score, and they are designed to be ``normalized". In particular, the mean is about 0, and the variance is approximately 1. Its magnitude is constructed to indicate the ``likelihood" of each event. One can set a simple threshold of 3 (the 3-sigma rule), and it will work if the z-score distribution is close to normal (Gaussian). However, for some sensory time series, the z-score distribution is far from normal. 

To make better use of the z-score, we set the two thresholds to be
\begin{equation}
    \beta^{-} = \min(-\gamma, Q_{\frac{\alpha}{2}}),
    \beta^{+} = \max(\gamma, Q_{1-\frac{\alpha}{2}}),
\end{equation}
where $\gamma$ is a threshold, and $Q_q$ denotes the q-quantile. The q-quantile is defined as the value of the quantile function (the inverse function of the cumulative distribution function) evaluated at $q$, where $q \in [0, 1]$. An event occurs if the z-score is below $\beta^-$ or above $\beta^+$. This is equivalent to checking two conditions. The first condition requires the absolute value of z-score exceeds a fixed threshold of $\gamma$ (usually taken to be 3). The second condition is based on the priori knowledge that events are sparse and occur with probability at most $\alpha$ (e.g., 0.01). The z-score must be in the lower or the upper $\alpha/2$ quantile. These two quantile thresholds are adaptive (data dependent), and robust in the case of non-normal data. In summary, this hybrid approach will be adaptive, simple, and robust.

\subsection{Graph Database for Sensor Event}
After event detection, the raw time series will be converted to tuples \textit{(event, sensor, time)}, which is interpreted as ``an event happens on a sensor at the time". If we represent this tuple with a graph, we need \textit{Event} node, \textit{Sensor} node, and \textit{Time} node. Each event tuple will be represented by these three nodes connected to each other. To store the events, we use a graph database \cite{angles2008survey}. A graph database is a database that uses graph structures for semantic queries with nodes, edges, and properties to represent and store data. The use of the graph database mainly consists of two phases: injecting data (nodes and relationships), and querying information.

\begin{figure}[th]
\centering
\includegraphics[width=0.6\columnwidth]{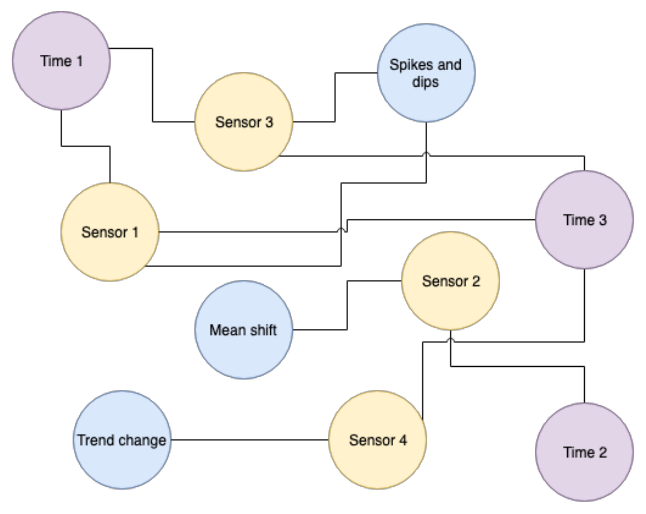} 
\caption{Structure of the graph database. There are three types of nodes: \textit{Event, Sensor, Time}. Each event will be represented by three connected nodes (event, sensor, time). Each node has its own properties. For example, \textit{Event} node contains event type, \textit{Sensor} node stores sensor information, and \textit{Time} node has time stamp.}
\label{fig:gdb}
\end{figure}

A sketch of the graph database structure is shown in Figure \ref{fig:gdb}. The graph database used in this study is neo4j \cite{neo4j}. The graph database will be dynamically updated in real-time. 

\subsection{Sensor Connection}
The next step will be querying this graph database and finding the connection between sensors. Intuitively, if two sensors have lots of events in common, they will have high connection weights.
 
A precise mathematical definition is needed to quantify this. Given a sensor $i$, denote all its event tuples within a given time window (say recent one month) as a set
\begin{equation}
    S_i = \{(e_k, t_k)\},
\end{equation}
where $e_k, t_k$ are the event type and event time respectively. $k$ is the index of the events ($k=1,2,3,\cdots$). Consider two sensors $(i, j)$, let their event set be $S_i = \{(e_k, t_k)\}$ and $S_j = \{(e_l, t_l)\}$ respectively. Fix a max time lag $\delta$, and define the concurrent event set of $j$ with respect to $i$ as follows:
\begin{equation}
    S(i,j) = \{(e_k, t_k) \ |\ \exists (e_l, t_l), s.t., e_l=e_k, |t_l-t_k| \leq \delta\}.
\end{equation}
Next, denote the concurrent event count as $C(i, j) = \textnormal{card}(S(i, j))$, where $\textnormal{card}(\cdot)$ is the cardinality of a set. $C(i,j)$ is the number of events of sensor $i$ for which sensor $j$ also has an event of the same type at about the same time (max time difference $\delta$). Intuitively, if $C(i,j)$ is large, then sensor $j$ has strong connection with sensor $i$.

Note that $C(i,j)$ is not symmetric, i.e., $C(i,j) \neq C(j,i)$ in general. Because the definition is asymmetric: we are standing at the viewpoint of sensor $i$. By definition, $C(i,i)=\textnormal{card}(S_i)$, i.e., the number of events of sensor $i$. Also, $C(i,j) \leq C(i,i)$, since $C(i,j)$ only counts a subset of $S_i$.

\begin{figure}[th]
\centering
\includegraphics[width=0.6\columnwidth]{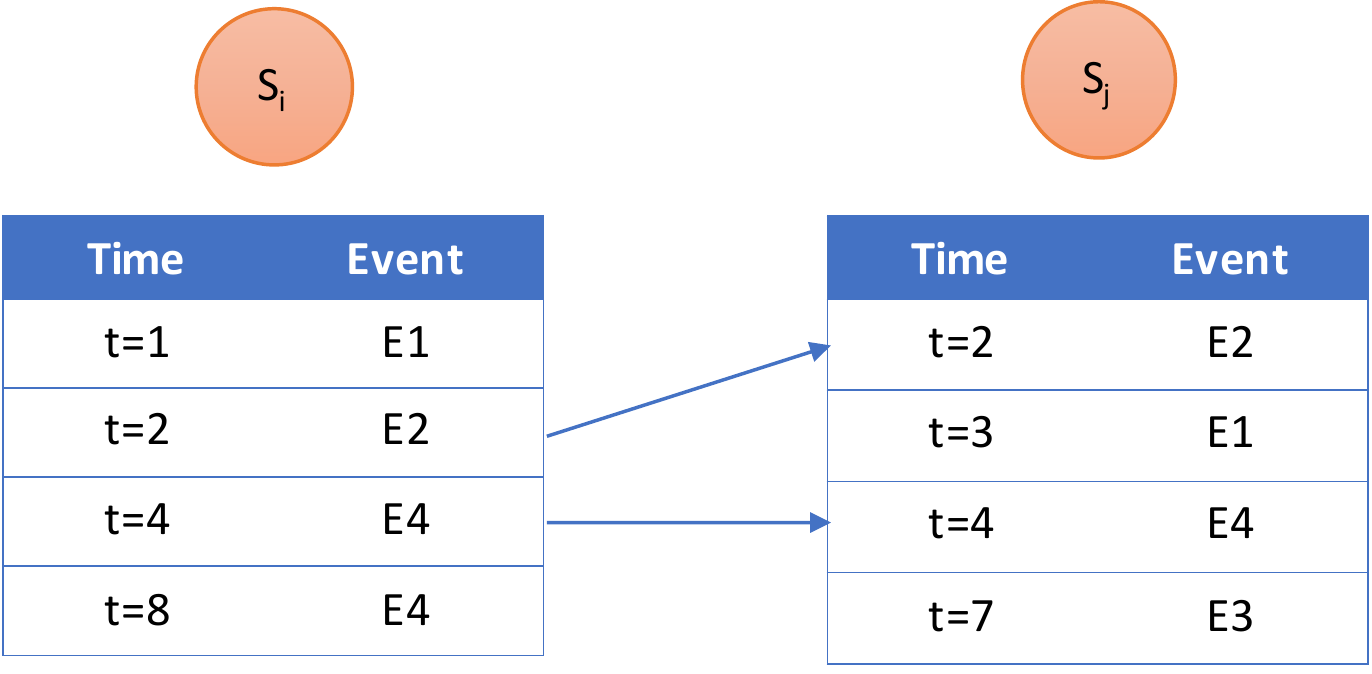} 
\caption{Sensor connection computation. Four types of events are considered: spike and dip (E1), mean shift (E2), variance shift (E3), and trend change (E4). The time window is $t \in [0, 10]$, and the max time lag is $\delta=0$. By definition, we have $C(i,j)=2$, $C(i,i)=4$, $P(i,j)=0.5$, $A(i,j)=0.25$.}
\label{fig:connection}
\end{figure}

The definition of concurrent count $C(i,j)$ is not normalized, i.e., it has different magnitudes for different sensors or different time windows. For a normalized measure, we define the probability of a concurrent event of sensor $j$ with respect to sensor $i$ as
\begin{equation}
    P(i,j) = C(i,j) / C(i,i),
\end{equation}
which is in the range $[0, 1]$. This quantity allows us to make meaningful comparison across different times and sensors. Finally, to make the sensor connection measure symmetric, we define the connectivity as
\begin{equation}
    A(i,j) = P(i,j) P(j, i),
\end{equation}
which is in the range $[0, 1]$. An example of sensor connection computation is shown in Figure \ref{fig:connection}.

\subsection{Sensor Graph}
Given any two sensors, a time window, and a max time lag, we will be able to compute their connection $A(i,j)$. This pairwise quantity provides insight into which sensors are more related than others. Besides, we want to view all the sensors as a whole unified object (graph). 

To begin with, we will define the notion of graph mathematically. A graph is \small$G=(A,X)$\normalsize, where \small$A$\normalsize\ is the adjacency matrix, and \small$X$\normalsize\ is the feature matrix. \small$A \in \mathcal{R}^{N \times N}$\normalsize, where $N$ is the number of nodes. \small$A(i,j)$\normalsize \ is the connection between two nodes $i,j$. \small$X \in \mathcal{R}^{N \times F}$\normalsize, where $F$ is the number of features for each node. $X(i,:)$ is the feature vector of node $i$.

The nodes and edges information are both contained in the adjacency matrix $A$. Two nodes $i,j$ are connected if $A(i,j) \neq 0$. In this work, $A$ will be symmetric, so the graph is undirected. In addition, $A(i,j)$ can take any values in $[0,1]$, so it is a weighted graph.

Consider multiple sensors, and if we put $A(i,j)$ into a matrix, then we will get a symmetric square matrix. This matrix will be naturally defined as the adjacency matrix of the sensor graph. For the feature matrix $X$, we will use the four features (generalized z-scores) defined for event detection.

\begin{figure}[th]
\centering
\includegraphics[width=0.5\columnwidth]{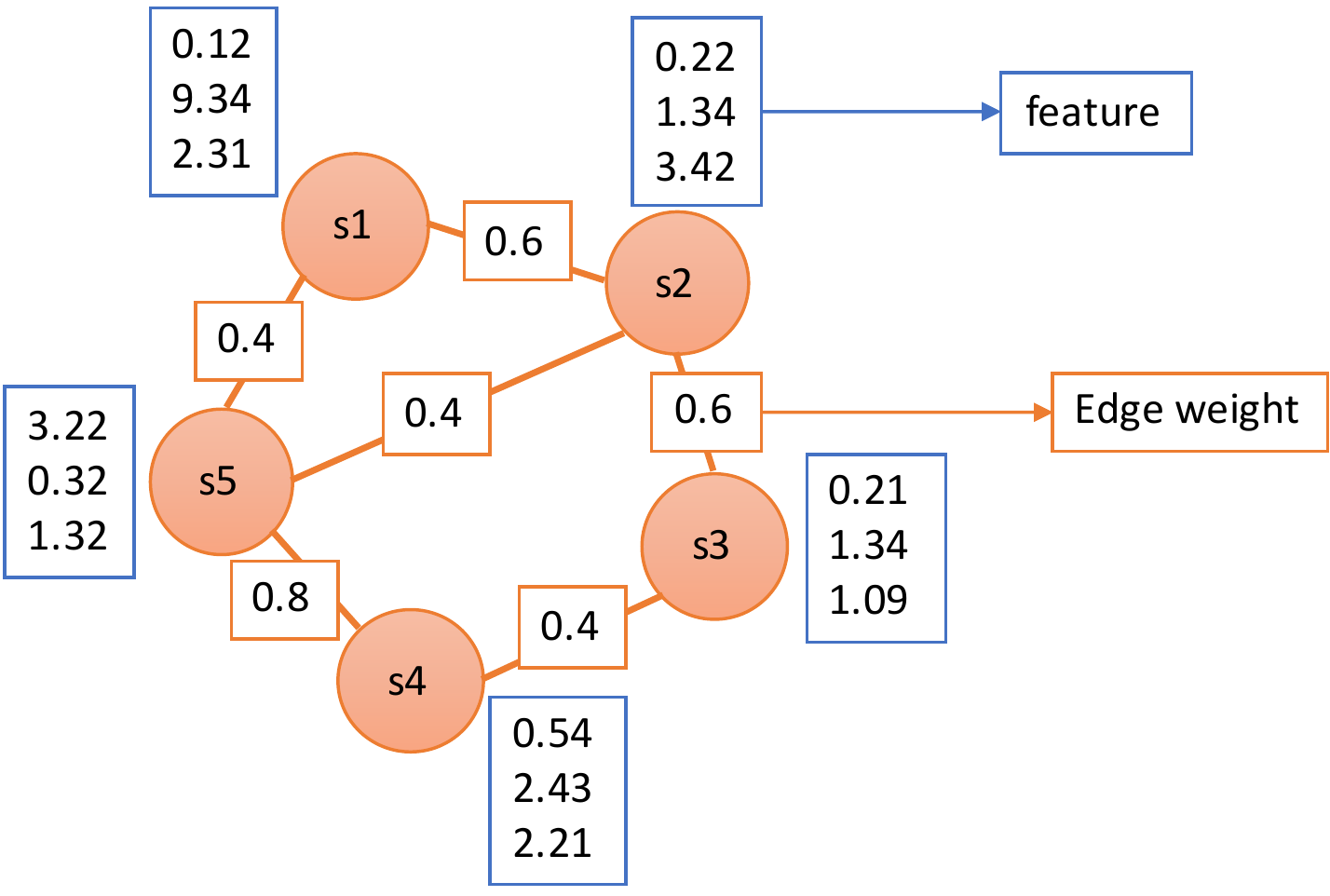}
\caption{A sketch of the spatial-temporal sensor graph $G(t)$ at a particular time instant.}
\label{fig:sensorgraph}
\end{figure}

Fix a time $t$, if we use the past events (say past one month) to compute $A(t),X(t)$, we will have a graph $G(t)$ that depends on time $t$. The sensor graph will be dynamic, and the dynamic graph is often referred to as spatial-temporal graph \cite{wu2019comprehensive}. The dynamic sensor graph is a powerful representation of the DC sensor time series. The graph can be constructed in real-time and analyzed online. A sketch of the sensor graph is shown in Figure~\ref{fig:sensorgraph}.

The sensor graph can be potentially used in many ways. For example, if an alarm occurs, the sensor graph can provide the most connected sensors related to the alarm. This will assist in root cause analysis \cite{rooney2004root}. Another typical use case is to reduce the false alarm rate in the DC monitoring system. Current anomaly detection systems are mainly based on univariate time series. A single sensor can be misleading and produces false alarms. For example, a single sensor might have a sudden jitter (due to wrong reading) and will trigger a (false) alarm. It is hard to tell if this is a false alarm by only looking at this single sensor. However, if the sensor graph is available, we can reference other related sensors. If all other related sensors are normal, then it is possible to filter out this false alarm.

\section{Graph of Things Prototype and Validation}
\label{sec:validation}
This section demonstrates the use of the above \textit{ts2graph} pipeline. We use real DC sensory data to build a DC graph of things (sensor graph). Next, we will validate this graph of things using two priori knowledge.

\subsection{Data Center Graph of Things Prototype}
We use the sensory data for a data center building at Alibaba Cloud. The sensors include IT load, real-time power usage effectiveness (PUE), outside environment sensors (6 sensors), water cooling system sensors (25 sensors), hot aisle temperature sensors (58 sensors), and computer room air conditioning (CRAC) unit fan speeds (72 sensors), for a total of 163 sensors. We follow the \textit{ts2graph} pipeline, and construct the sensor graph. The time series measurement frequency is five minutes. The recent one month data (May 12, 2018 to June 12, 2018) is used, and the max time lag is fifteen minutes.

\begin{figure}[th]
\centering
\includegraphics[width=0.8\columnwidth]{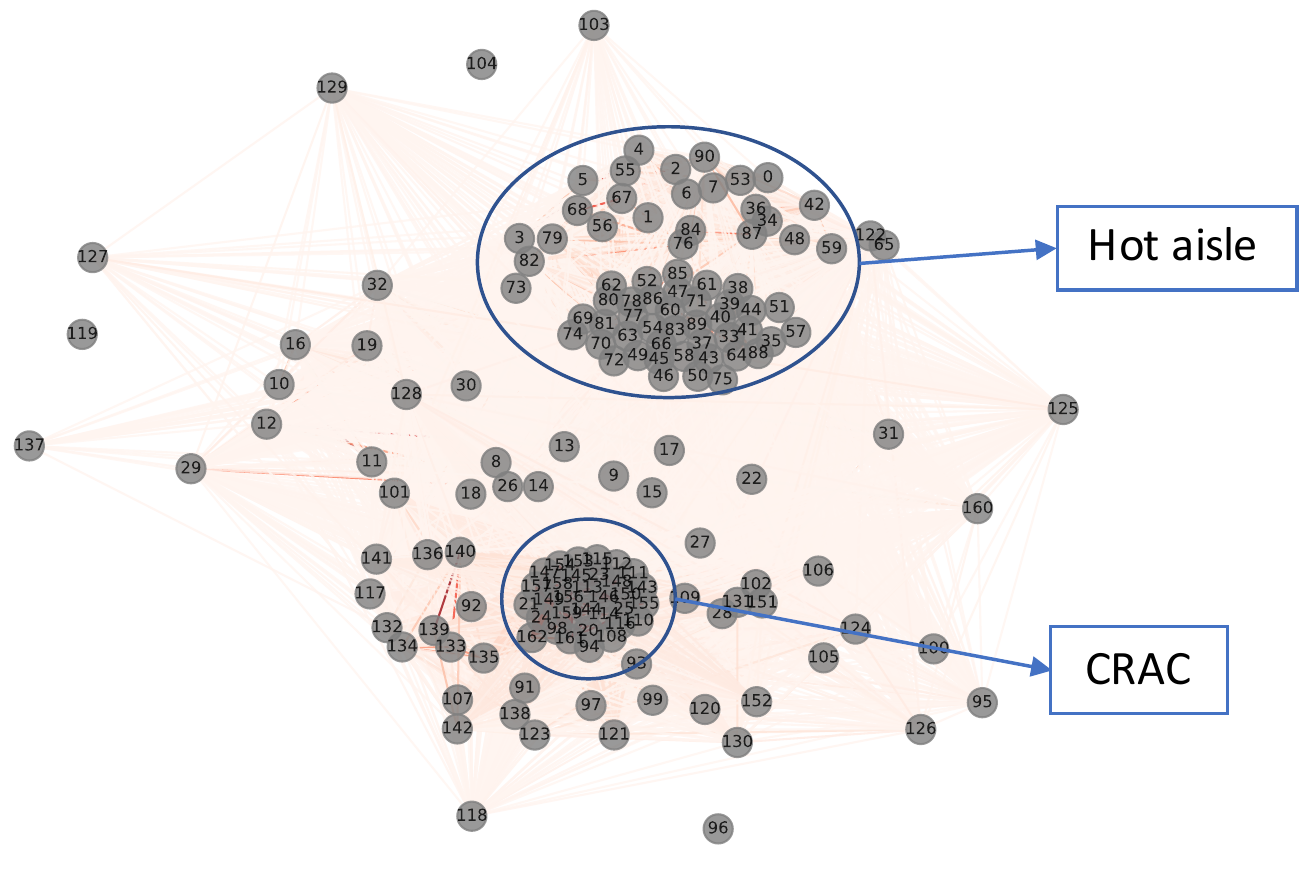} 
\caption{A snapshot of the sensor graph. Nodes with high connections are closer, and edge lightness indicates connection magnitude. There are two clusters, and they correspond to hot aisles and CRACs.}
\label{fig:dcgraph}
\end{figure}

A snapshot of the sensor graph is depicted in Figure \ref{fig:dcgraph}. First of all, there are two clusters, and they correspond to hot aisles and CRACs. The cluster structures are consistent with our priori knowledge. This gives us confidence in \textit{ts2graph}.

\begin{figure}[th]
\centering
\includegraphics[width=0.7\columnwidth]{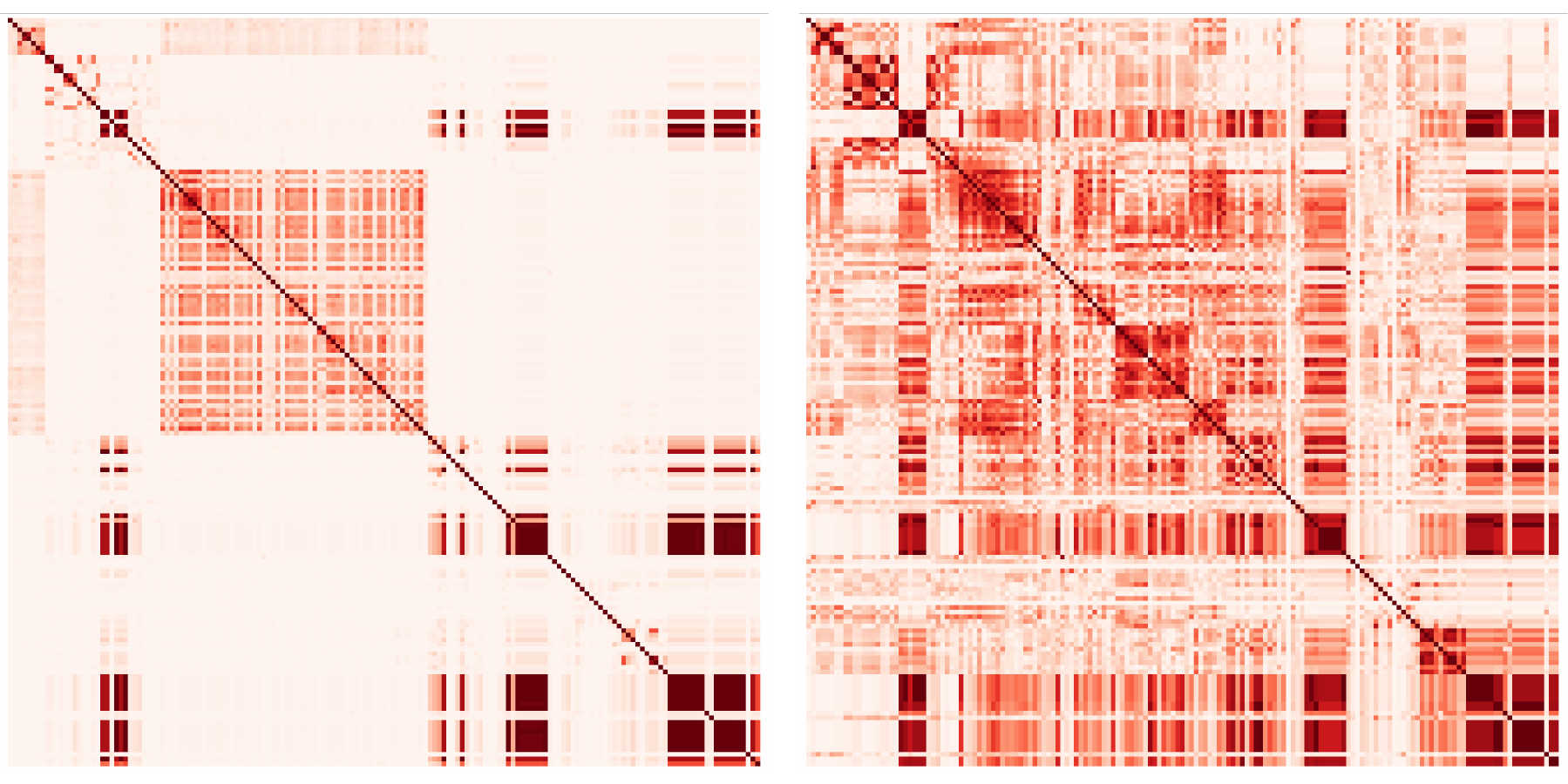} 
\caption{Heatmap of the adjacency matrix (left) and Pearson correlation matrix (right). The adjacency matrix is more ``sparse", in fact half of the values are less than 0.01.}
\label{fig:adj_corr}
\end{figure}

Also, we compare the adjacency matrix with the Pearson correlation matrix, as displayed in Figure \ref{fig:adj_corr}. Pearson correlation matrix is computed from the same one month data, and we take its absolute value. Each entry of the Pearson correlation matrix is defined as the Pearson correlation coefficient between the two time series. It is visible that the adjacency matrix is much more ``sparse". The reason is that events are much more sparse in time and reveal the true behaviors of the sensors. However, the raw time series may contain lots of noise, and the direct correlation coefficient may be misleading and indicates spurious correlations. This characteristic is very appealing. A typical application will be finding out which sensors have a strong connection with a target sensor. We can identify them by merely looking at a particular row of the matrix. After sorting the connection weight in descending order, we then pick the top connected sensors. Since the adjacency matrix is more ``sparse", we will be more confident about which sensor is related (or unrelated) to the target sensor.

\subsection{Validation with Priori Knowledge}
Given the constructed DC graph of things prototype, we want to validate further that it reveals meaningful relationships. To verify the sensor graph, we must have some ground truth to compare against. The ground truth could come from a priori knowledge. For example, suppose that we know two sensors are physically close to each other, and measure the same quantity (e.g., temperature). Then they should have high connection weight. The adjacency matrix of the sensor graph should support this fact. For validating the sensor graph (in particular the adjacency matrix), we will make use of two priori knowledge.

\subsubsection{Related Outside Environment Sensors}
First, the outside environment sensors include three types of sensors: temperature, humidity, and wet-bulb temperature (WBT). The technical definition of WBT \cite{gupton2001hvac} is unnecessary here, and can be simplified as: WBT is determined by both temperature and humidity. Besides, each of the three sensors has a redundant sensor. The redundant pair should have a very high connection weight. In summary, these six sensors are strongly connected. This priori knowledge is used as the ground truth. We design the following experiment to verify if the adjacency matrix (or correlation matrix) can recover these relationships:

\begin{enumerate}
\setlength\itemsep{-0.2em}
\item pick a matrix (adjacency matrix or any of the three correlation matrices)
\item choose a target sensor from the six outside environment sensors and get its connection (with all other sensors) from the row of the matrix corresponding to that sensor
\item sort connection in descending order, and pick the top six related sensors
\item count the number of sensors in the top six list that comes from the six outside environment sensors (the priori knowledge), divide by six, and call the result $m_1$
\item sum up the connection weight of sensors that are in both the top six list and six outside environment sensors list, divide by the total connection weight of target sensor (i.e., the sum of the matrix row), and denote this quantity as $m_2$
\item sum up the connection weight of the six outside environment sensors, and divide by the total connection weight of the target sensor, let this number be $m_3$
\item repeat step 2-6 and average $m_1, m_2, m_3$ across all six sensors
\end{enumerate}

The evaluation metric $m_1, m_2, m_3$ are designed such that the higher, the better. For example, $m_1$ is roughly the probability of recovering the top connected sensors (defined by the priori knowledge) from the top connection list (encoded by the matrix). $m_2$ measures the ``confidence" when recovering the top connected sensors (the priori knowledge). $m_3$ measures the total connection percentage of the top connected sensors (the priori knowledge). They are all normalized to be in the range $[0, 1]$.

For comparison, we also conduct the same experiment on the Pearson correlation matrix, Spearman rank correlation \cite{spearman1904proof} matrix, and Kendall rank correlation \cite{kendall1938new} matrix. These three correlation matrices are all computed from the same one month data. Also, we take absolute value for the correlation matrices so that its value is in $[0, 1]$. Each entry of the correlation matrix is defined as the correlation coefficient of the two time series. The experiment results are reported in Table~\ref{tab:validation1}. The adjacency matrix achieves the best result among all the matrices. Notice that $m_1=1.00$ (100\%) for the adjacency matrix, which means that the adjacency matrix always finds the six related sensors (the priori knowledge) in its top six list (from the matrix). In terms of $m_2, m_3$, adjacency matrix outperforms correlation matrices by a factor of 2.

\begin{table}[th]
\smallskip
\centering
\resizebox{.55\columnwidth}{!}{
\smallskip
\begin{tabular}{|c||c|c|c|c|}
\hline
        & Adjacency & Pearson & Spearman & Kendall\\ \hline \hline
$m_1$ & 1.00 & 0.77 & 0.77 & 0.75 \\ \hline
$m_2$ & 0.35 & 0.14 & 0.13 & 0.15 \\ \hline
$m_3$ & 0.35 & 0.15 & 0.13 & 0.15 \\ \hline
\end{tabular}
}
\caption{Evaluation metric for validation experiment based on six related outside environment sensors. Adjacency matrix outperforms correlation matrices in terms of all evaluation metrics.}
\label{tab:validation1}
\end{table}

\subsubsection{Related Temperature Sensors in Computer Room}
Second, hot aisle temperature sensors from the same computer room should have strong connections (the priori knowledge). Because they are physically close to each other, and both are measuring the same physical quantity (temperature). There are seven computer rooms in the DC used in this work, each with eight hot aisle temperature sensors. Therefore, we can carry out a similar experiment as before. This time, we replace the top-six list by the top-eight list, and the priori knowledge of strongly connected sensors will be the sensors from the same computer room. The evaluation metric will be averaged for all hot aisle sensors. The results are reported in Table \ref{tab:validation2}. Again, graph adjacency matrix beats correlation matrices for all metrics. For $m_2, m_3$ it outperforms by about 50\%.

\begin{table}[th]
\smallskip
\centering
\resizebox{.55\columnwidth}{!}{
\smallskip
\begin{tabular}{|c||c|c|c|c|}
\hline
        & Adjacency & Pearson & Spearman & Kendall\\ \hline \hline
$m_1$ & 0.46 & 0.43 & 0.45 & 0.46 \\ \hline
$m_2$ & 0.16 & 0.06 & 0.09 & 0.11 \\ \hline
$m_3$ & 0.22 & 0.11 & 0.14 & 0.15 \\ \hline
\end{tabular}
}
\caption{Evaluation metric for validation experiment based on related hot aisle temperature sensors. Adjacency matrix achieves best performance in all evaluation metrics.}
\label{tab:validation2}
\end{table}

\section{Graph of Things for Anomaly Detection}
\label{sec:dcanomaly}
In the previous section, the sensor graph has been validated using two priori knowledge. In this section, we will demonstrate the use of the sensor graph with anomaly detection.

\subsection{Anomaly Detection Methods}

\subsubsection{Anomaly Detection}
Data center (DC) is a complex system with lots of components, which includes IT devices and supporting equipment. These components are interrelated, and they form the internet of things (IoT). Anomaly detection is crucial for the healthy running of DC.

Traditional anomaly detection mainly deals with time series, or vector inputs \cite{chandola2009anomaly}. Recently, there have been some deep learning based anomaly detection methods \cite{chalapathy2019deep}. However, the inputs of these algorithms are mainly vectors. These methods will work if the underlying data can be well described by vectors.

However, in DC, IoT data sources are related in complex ways. The graph models both the features of time series (node feature matrix) and their connections (adjacency matrix), which makes it a good candidate as the knowledge representation between DC sensors.

There have been some graph-based anomaly detection methods proposed in recent years \cite{akoglu2015graph}. However, these algorithms are still mainly based on conventional graph theory, such as the shortest path and connected components. Though they process more information than the vector-based methods, those methods are still ``shallow". If the underlying graph data input has complex structures or hidden information, it will be hard for these methods to work effectively. 

\subsubsection{Graph Neural Network}
To address this challenge, we can combine deep learning, graph as knowledge representation, and anomaly detection. Deep learning is able to discover hidden information and structure from data \cite{lecun2015deep}, and the input is usually a vector or image. Now the input is a graph. Therefore we use deep learning for the graph, which is referred to as graph neural network (GNN) \cite{akoglu2015graph,zhou2018graph}. GNN extends the application domain of deep learning from vector or image to a graph. A graph is a versatile data structure with strong representation power. For example, an image can be viewed as a simple graph where each pixel is a node, and it is only connected to 8 nearby pixels.

Convolutional neural network (CNN) operates on image \cite{lecun1995convolutional}. The most fundamental operation of CNN is image convolution. In short, image convolution aggregates information from nearby pixels for feature extraction \cite{dumoulin2016guide}. 

GNN generalizes CNN to work on the graph \cite{scarselli2008graph}. The basic building block of GNN is graph convolution, where the information of connected nodes is combined to extract features for each node. The main difference from image convolution is that nearby pixels are replaced by connected nodes in the graph.


\subsubsection{Methodology}
To demonstrate the use of the sensor graph, we apply GNN to the sensor graph for the purpose of anomaly detection. In particular, we will apply the graph auto-encoder (GAE) \cite{wu2019comprehensive}. As benchmark, we also consider conventional vector auto-encoder (VAE) \cite{goodfellow2016deep} and principal component analysis (PCA) \cite{pearson1901liii}. GAE compresses graph into hidden states (with lower dimension), and reconstructs it back. GAE is able to learn the hidden low dimensional structure in graphs directly. VAE learns the nonlinear low dimensional subspace on vector inputs. PCA works by finding a linear subspace (of vectors) that best explains the variance in data.

Both these three methods fall into the category of subspace-based methods for anomaly detection \cite{chandola2009anomaly}. Subspace based method assumes that most data are ``normal" and lie in some low dimensional subspace (linear or nonlinear). It tries to learn this subspace from data, then project data onto the subspace, and finally reconstruct the original input. ``normal" data will have small reconstruction error, but ``anomaly" data will have a higher reconstruction error. Therefore, one can differentiate ``anomaly" from ``normal" based on the reconstruction error.

There exist many variants of graph convolution, as summarized in \cite{wu2019comprehensive}. Some are based on graph spectral theory, while others are spatially based. In this work, we define the following graph convolution operation:
\begin{equation}
\small
    X_{k+1} = \textnormal{GCN}(A, X_k) = U\hat{A} X_k W + B,
\end{equation}
\normalsize
where \small $A \in \mathcal{R}^{N \times N}$ \normalsize is the adjacency matrix, \small $X_k \in \mathcal{R}^{N \times F_k}$ \normalsize is the feature matrix at layer $k$. $N$ is the number of nodes, and $F_k$ is the number of features for each node. $\hat{A}$ is the normalized adjacency matrix, defined as \small$\hat{A} = D^{-\frac{1}{2}} A D^{-\frac{1}{2}}$\normalsize, where $D$ is a diagonal matrix of node degrees, \small$D(i,i)=\sum_j A(i,j)$\normalsize. \small$U \in \mathcal{R}^{N \times N}$, $W \in \mathcal{R}^{F_k \times F_{k+1}}$\normalsize are parameters, and \small$B \in \mathcal{R}^{N \times F_{k+1}}$\normalsize is the bias. Notice that graph adjacency matrix is kept fixed across GCN layers, while the feature matrix is transformed. The transformation of feature matrix depends on adjacency matrix, i.e., the topology of the graph. After convolution, a tanh activation function is be applied.

The GAE used here consists of 3 layers. The first layer reduces the number of features per node from 4 to 3, and applies the tanh activation. The second layer reconstructs the number of features from 3 to 4, followed by the tanh transformation. The third layer is a rescaling layer, where only graph convolution is applied, without activation function. The target output of GAE is the original feature matrix, and backpropagation \cite{goodfellow2016deep} is used to update the parameters.

For VAE and PCA, the input is the feature vector of the whole graph. The feature vector of the whole graph is obtained by flattening the feature matrix \small$X \in \mathcal{R}^{N \times 4}$\normalsize. As a result, the feature vector has dimension \small$\mathcal{R}^{4N}$\normalsize. VAE has the same three-layer structure as GAE, and the number of hidden states is equal to $3N$. For PCA, we keep the top $3N$ principle components in the subspace, and reconstruction is based on $3N$ principal components. The characteristics of the three described methods are summarized in Table~\ref{tab:algo}.

\begin{table}[th]
\smallskip
\centering
\resizebox{.99\columnwidth}{!}{
\smallskip
\begin{tabular}{|c||c|c|c|c|}
\hline
      & Input & Subspace & Hidden state & Output (target)
        \\ \hline \hline
Graph auto-encoder (GAE) & $G=(A,X)$ & Nonlinear & $X_h \in \mathcal{R}^{N \times 3}$ & $G=(A,X)$\\ \hline
Vector auto-encoder (VAE) & $x = \textnormal{vec}(X)$ & Nonlinear & $x_h \in \mathcal{R}^{3N}$ & $x = \textnormal{vec}(X)$\\ \hline
Principal component analysis (PCA) & $x = \textnormal{vec}(X)$ & Linear & $x_h \in \mathcal{R}^{3N}$ & $x = \textnormal{vec}(X)$\\ \hline
\end{tabular}
}
\caption{Summary of algorithms used. Both algorithms are replicating the input as the output. $\textnormal{vec}(X)$ denotes the flattened vector of a the feature matrix $X$.}
\label{tab:algo}
\end{table}

\subsection{Synthetic Anomaly Generation}
The DC sensory data does not come with human expert anomaly labels. To circumvent this issue, we choose to inject synthetic anomaly to the real data. In this work, we focus on the anomaly of a related group of sensors. Group anomaly is harder to detect and is of greater importance (corresponding to severe problems). For example, if the hot aisle temperature sensors from the same computer room are experiencing rapid temperature increase simultaneously, this is definitely an anomaly. In contrast, if a single sensor has a sudden short-time jitter (due to false reading), and all other related sensors stay the same, then this is probably not considered an anomaly. It is very challenging for traditional time series based anomaly detection methods to detect group anomaly. We consider two ways to generate synthetic anomaly. In the first method, we modify the feature matrix directly. This is a necessary check for GNN to verify that it works for anomaly graphs. In the second approach, we modify the original time series directly. This is more realistic because real anomaly originates from the raw time series.

\subsubsection{Feature Matrix Anomaly}
First, we add anomaly to the feature matrix of sensors directly. Note that the features (generalized z-scores) are designed to be ``normalized", i.e., the mean is about 0 and variance is about 1. At each time step, we set a 20\% probability to modify the feature matrix. For each modification, we randomly choose a computer room (7 of them), and select the group of hot aisle temperature sensors from this computer room (8 sensors). Then we add (or subtract) random perturbation drawn from normal distribution $N(3, 1)$ to the features corresponding to these sensors. The anomaly label is set according to the affected sensors. Anomaly label will have the same dimension as the number of sensors. The anomaly label is of ``sensor" level, i.e., it gives a label to each sensor. Of course, both feature matrix and anomaly label will be dependent on time. As a result, we have an anomaly label for each sensor at each time step.

\subsubsection{Time Series Anomaly}
A more realistic approach to generate anomaly is to modify the original time series directly. As proposed, we care more about the anomaly of a related subgroup. To simulate this scenario, we design the following anomaly generation. At each time step, there is a 20\% probability of modifying the time series. Then, we pick a computer room at random and find the group of hot aisle temperature sensors in this computer room (8 sensors). For this sensor group, we add perturbation drawn from normal distribution $N(6, 1)$ (unit: Celsius degree) to the next one hour time series. The anomaly label is assigned according to which sensor has an anomaly, and lasts for one hour. Therefore, the anomaly label is of ``sensor" level and marks anomaly sensor for each time step.

\subsection{Experiment Result}
In this section, we will show the result for two types of synthetic anomaly data. The training dataset contains four-week original clean data. The validation dataset contains one-week original data. The testing dataset contains one-week synthetic anomaly data. The measurement frequency of the time series is five minutes.

\subsubsection{Evaluation metric}
As an anomaly detection task, we use precision, recall and F1-score \cite{powers2011evaluation} to evaluate the algorithm performance. Precision is defined as
\begin{equation}
    \textnormal{precision} = \frac{\textnormal{tp}}{\textnormal{tp} + \textnormal{fp}},
\end{equation}
where tp is the number of true positive, and fp is the number of false positive. Recall is defined as
\begin{equation}
    \textnormal{recall} = \frac{\textnormal{tp}}{\textnormal{tp} + \textnormal{fn}},
\end{equation}
where fn is the number of false negative cases. Intuitively, precision is the precision of an anomaly alarm (whether a predicted anomaly is an actual anomaly). Recall measures the ability to find all anomalies (how many anomalies are found). F1-score combines precision and recall into a single number, and is defined as the harmonic mean of precision and recall.

\subsubsection{Evaluation Result}
The anomaly score (``likelihood" of an anomaly) is based on the reconstruction error. Reconstruction error has the same dimension as the feature matrix, but we have a ``sensor" level anomaly label. Therefore, we will aggregate the reconstruction error by taking the average across four features for each sensor. The anomaly score is defined to be this ``sensor" level reconstruction error. Based on the anomaly score, one can obtain a precision (F1 score)-recall curve by adjusting the threshold of anomaly prediction \cite{chandola2009anomaly}.

\begin{figure}[th]
    \centering
     \begin{subfigure}{0.4\textwidth}
         \centering
         \includegraphics[width=0.9\textwidth]{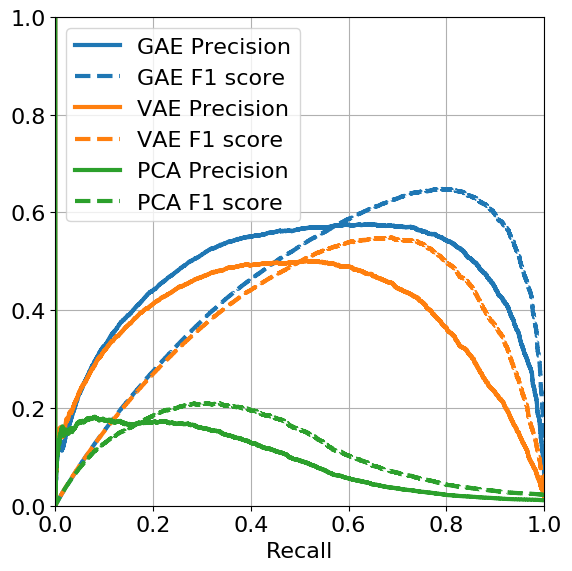}
         \caption{Feature matrix anomaly}
     \end{subfigure}
     \begin{subfigure}{0.4\textwidth}
         \centering
         \includegraphics[width=0.9\textwidth]{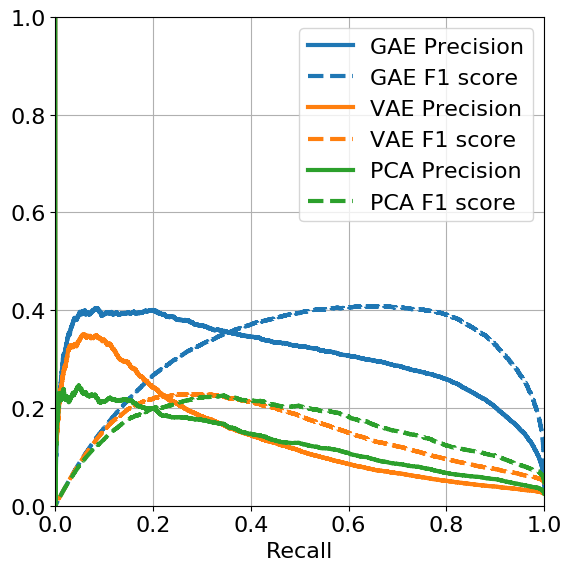}
         \caption{Time series anomaly}
     \end{subfigure}
    \caption{Precision (F1 score)-recall curve for anomaly detection on two synthetic dataset. GAE achieves the best preformance for both cases. }
    \label{fig:result}
\end{figure}

The results for the performance of graph auto-encoder (GAE), vector auto-encoder (VAE), and principal component analysis (PCA) are shown in Figure \ref{fig:result}. For both datasets, GAE achieves the best performance, because it takes into account the topology relationship between sensors. VAE outperforms PCA because the sensor relationship is nonlinear. For the first dataset, GAE outperforms PCA by a factor of 3 to 4 in terms of both precision and F1 score for a wide range of recall ([0.2, 0.8]). In the second dataset, GAE gets twice precision (and F1 score) compared to VAE and PCA. Note that the time series anomaly dataset is more challenging, as expected. For more realistic time series anomaly, VAE performance is no longer close to GAE.

\section{Conclusion}
In this work, we propose \textit{ts2graph}, a data-driven pipeline for constructing a DC graph of things based on sensory time series. A prototype sensor graph (i.e., graph of things) is built with a real DC sensor time series dataset, and it is validated using two priori knowledge. The sensor graph is capable of revealing meaningful relationships between sensors. To demonstrate the use of the sensor graph for anomaly detection, we evaluate the performance of the graph neural network. In particular, we apply graph auto-encoder (GAE) for anomaly detection and compare against classical methods, including vector auto-encoder (VAE) and principal component analysis (PCA). GAE stands out by a factor of 2 to 3 (in terms of precision and F1 score), because it takes into account the topology relationship between sensors. The sensor relationships can not be captured by vector-based methods like VAE and PCA, and have to be described by a graph. Based on this study, we expect the DC graph of things can serve as the infrastructure for the DC monitoring system. DC graph of things will pave the way for both anomaly detection and root cause analysis.

\section*{Acknowledgments}
\label{sect:acks}
The authors kindly acknowledge Yue Li (Alibaba Group Inc) for helpful discussions about data and proofreading the paper.

\label{sect:bib}
\bibliographystyle{plain}
\bibliography{ref}



\end{document}